\documentclass[conference,letterpaper]{IEEEtran}
\IEEEoverridecommandlockouts
\usepackage[dvips]{graphicx}  

\usepackage{multirow}
\usepackage[left=0.71in,top=0.94in,right=0.71in,bottom=1.18in]{geometry}
\usepackage{cite}
\usepackage{color}
\usepackage{fullpage} 
\usepackage{float} 
\usepackage{caption}
\usepackage{subcaption}
\usepackage{amsmath}
\usepackage{amssymb}
\usepackage{booktabs}
\usepackage{adjustbox}
\usepackage{tabularx}
\usepackage{multirow}
\usepackage{gensymb}
\usepackage{soul}
\usepackage[textsize=tiny,textwidth=2cm]{todonotes}
\newcommand{\lorenzo}[1]{\textcolor{black}{#1}}

\graphicspath{{figures/}}

\newcommand{\tact}{$\mathbf{\tau}$}
\newcommand{\graspInit}{$\mathbf{\Theta_{grasp}^{init}}$}
\newcommand{\graspFin}{$\mathbf{\Theta_{grasp}^{fin}}$}
\newcommand{\wrap}{$\mathbf{\Theta_{wrap}}$}
\newcommand{\thetaGrasp}{$\mathbf{\Theta_{grasp}}$}
\newcommand{\thetaAll}{$\mathbf{\Theta_{all}}$}

\begin{document}
\title{\huge \lorenzo{Controlled Tactile Exploration and Haptic Object Recognition}\footnoterule\thanks{This research has received funding from the European Union's Seventh Framework Programme for research, technological development and demonstration under grant agreement No. 610967 (TACMAN).}}

\author{\authorblockN{ Massimo Regoli, Nawid Jamali, Giorgio Metta and Lorenzo Natale}
\authorblockA{\textit{iCub Facility}\\
\textit{Istituto Italiano di Tecnologia}\\
\textit{via Morego, 30, 16163 Genova, Italy}\\
\textit{\{massimo.regoli, nawid.jamali, giorgio.metta, lorenzo.natale\}@iit.it}\\}%
}%

\maketitle
\begin{abstract}
In this paper we propose a novel method for in-hand object recognition. The method is composed of a grasp stabilization controller and two exploratory behaviours to capture the shape and the softness of an object. Grasp stabilization plays an important role in recognizing objects. First, it prevents the object from slipping and facilitates the exploration of the object. Second, reaching a stable and repeatable position adds \lorenzo{robustness to the learning algorithm and increases invariance with respect to the way in which the robot grasps the object}. The stable poses are estimated using a Gaussian mixture model (GMM).
We present experimental results showing that using our method the classifier can successfully distinguish 30 objects. We also compare our method with a benchmark experiment, in which the grasp stabilization is disabled. We show, with statistical significance,  that our method outperforms the benchmark method.

\end{abstract}


\section{Introduction}
Sense of touch is essential for humans. We use it constantly to interact with our environment. Even without vision, humans are capable of manipulating and recognizing objects. Our mastery of dexterous manipulation is attributed to well developed tactile sensing~\cite{Howe1994}. To give robots similar skills, researchers are studying use of tactile sensors to help robots interact with their environment using the sense of touch. Furthermore, different studies show the importance of tactile feedback when applied to object manipulation \cite{dahiya2010tactile}\cite{yussof2009grasp}.

Specifically, in the context of object recognition, tactile sensing provides information that cannot be acquired by vision. Indeed, properties such object texture and softness can be better investigated by actively interacting with the object. In order to detect such properties, different approaches have been proposed. Takamuku et al.~\cite{takamuku2007haptic} identify material properties by performing tapping and squeezing actions. Johansson and Balkenius~\cite{johnsson2008recognizing} use a hardness sensor to measure the compression of materials at a constant pressure, categorizing the objects as hard and soft. Psychologists have shown that humans make specific exploratory movements to get cutaneous information from the objects~\cite{lederman1987hand}, that include, pressure to determinate compliance, lateral sliding movements to determinate surface texture, and static contact to determine thermal properties. Hoelscher et al.~\cite{hoelscher2015evaluation} use these exploratory movements to identify objects based on their surface material, whereas other researchers have focused on how to exploit them to reduce the uncertainty in identifying object properties of interest~\cite{xu2013tactile}.


\begin{figure}[t]
\includegraphics[width=0.8\linewidth]{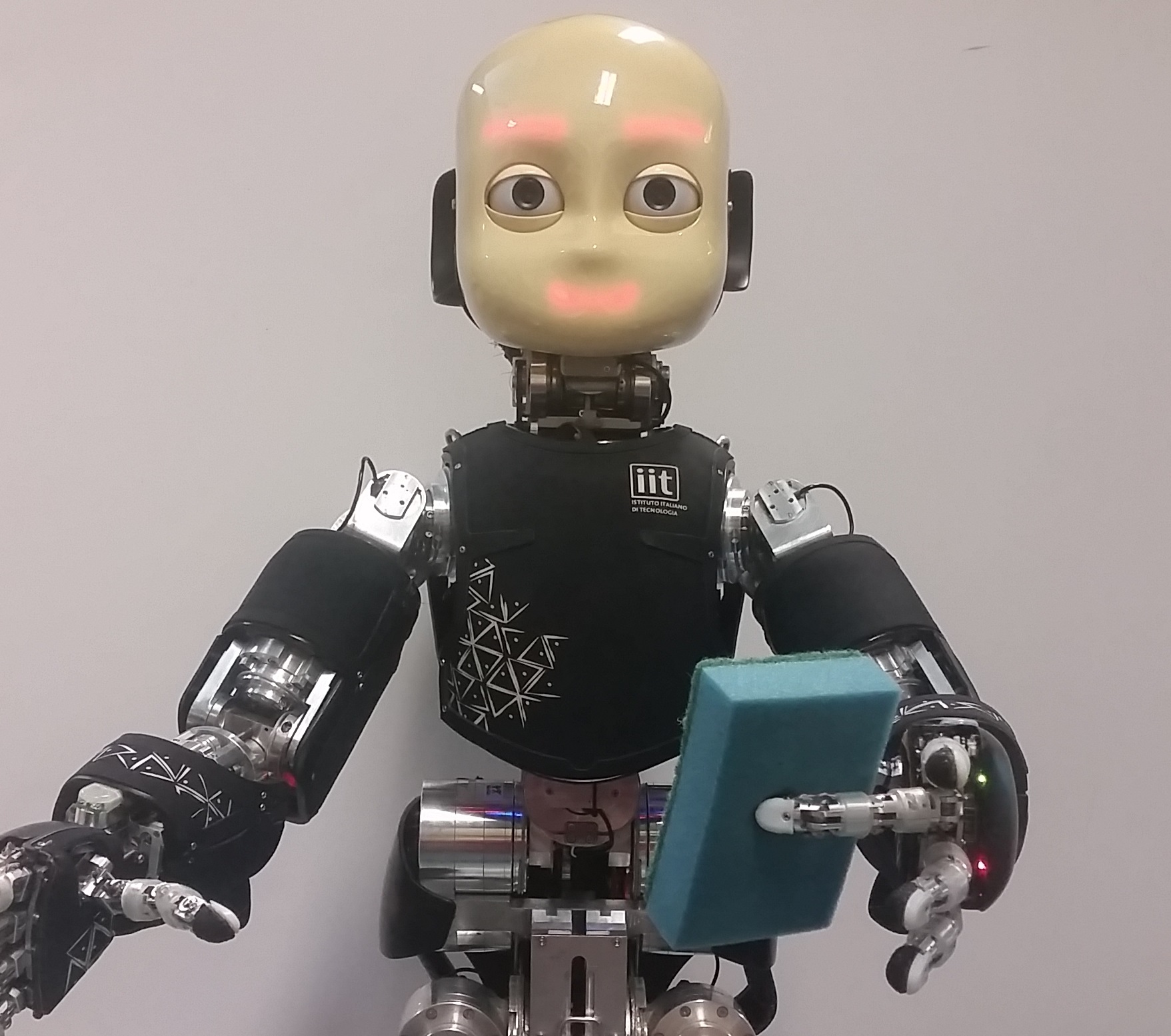}
    \centering
    \caption{The iCub humanoid robot carrying out the object recognition task.}
    \label{fig:icub}
\end{figure}

All these approaches carry out exploratory movements using a single finger and assume that the object does not move. Conversely, other works recognize an object by grasping the object, putting less restrictions on the hand-object interaction. Schneider et al.~\cite{schneider2009object} propose a method in which each object is grasped several times, learning a vocabulary from the tactile observations. The vocabulary is then used to generate a histogram codebook to identify the objects. Chitta et al.~\cite{chitta2010tactile} propose a method that, using features extracted while grasping and compressing the object, can infer if they are empty or full and open or close. Chu et al.~\cite{chu2013using} perform exploratory movements while grasping the object in order to find a relationship between the features extracted and haptic adjectives that humans typically use to describe objects.

However, most of these approaches do not deal with the stability problem and assume that the object is laying on, or are fixed to a surface such as a table. When the object has to be held in the robot's hand, stability problems such as preventing it from falling, make the task of extracting features through interactions more challenging. Kaboli et al.~\cite{kaboli2015hand} recognise objects using their surface texture by performing small sliding movements of the fingertip while holding the object in the robot's hand. Gorges et al.~\cite{gorges2010haptic} merge a sequence of grasps into a statistical description of the object that is used to classify the objects. In a recent work Higy et al.~\cite{higy2016combining} propose a method in which the robot identifies an object by carrying out different exploratory behaviours such as hand closure, and weighing and rotating the object. In their method the authors fuse sensor data from multiple sensors in a hierarchical classifier to differentiate objects.

In these approaches the stability is typically managed by performing a power grasp, that is,  wrapping all the fingers around the object. This means that in general, the final hand configuration after the grasp is not controlled. It strictly depends on the way the object is given to the robot. Due to this, the tactile and proprioceptive feedback suffer from high variability. \lorenzo{This requires a larger number of grasps to be performed and negatively affects the performance}. Moreover, performing power grasps may limit further actions that could help in extracting other object features such as softness/hardness.

In this work we propose a novel method for in-hand object recognition that uses a controller proposed by Regoli et al~\cite{regoli2016hierarchical} to stabilize a grasped object. The controller is used to reach a stable grasp and \lorenzo{reposition the object in a repeatable way}.
We perform two exploratory behaviours: squeezing to capture the softness/hardness of the object; and wrapping all of the fingers around the object to get information about its shape. The stable pose achieved is unique given the distance between the points of contact (related to the size of the object), resulting in high repeatability of features, which improves the classification accuracy of the learned models. Differently from other methods, we do not put any restrictions on the objects.


We validated our method on the iCub humanoid robot~\cite{metta2008icub} (Fig.~\ref{fig:icub}). We show that using our method we can distinguish 30 objects with \lorenzo{99.0\% $\pm$ 0.6\% accuracy}. We also present the results of a benchmark experiment in which the grasp stabilization is disabled. We show that the results achieved using our method outperforms the benchmark experiment.

In the next section we present our method for in-hand object recognition. In section \ref{sec:experiments} we describe the experiments carried out to validate our method, while in section \ref{sec:results} we present our results. Finally, in section \ref{sec:conclusions} we conclude the paper and provide future directions.

\section{Methodology}
\label{sec:methodology}

Here we present the method used to perform the in-hand object recognition task. We use an anthropomorphic hand, but the method can be easily extended to any type of hand that has at least two opposing fingers. \lorenzo{We use the tactile sensors on the fingertips of the hand~\cite{jamali2015new}, which provide pressure information on 12 taxels for each fingertip. An important assumption in this work is that the object is given to the robot by a collaborative operator, in such a way that the robot can grasp it by closing the fingers. The remaining steps are performed by the robot autonomously, namely:}
\begin{itemize}
\item grasping the object using a precision grasp, that is, using the tip of the thumb and the middle finger,
\item reaching an optimal stable pose,
\item squeezing the object to get information about its softness,
\item wrapping all the fingers around the object to get information about its shape.
\end{itemize}

We start by giving an overview of the grasp stabilizer component. This is followed by a description of the feature space, and then we give a brief overview of the machine learning algorithm used to discriminate the objects.

\subsection{Grasp stabilization}
\label{sec:stabilization}

Grasp stabilization is a crucial component of our method for two reasons. First, it is needed to prevent the object from falling, for example, when executing actions like squeezing. Second, reaching a stable and repeatable pose for a given object improves the classifier accuracy. We use our previously developed method to stabilize the object~\cite{regoli2016hierarchical}. In the rest of this section we quickly revise this method and explain how we apply it to our problem (details of the controller can be found in~\cite{regoli2016hierarchical}). In this paper we use two fingers instead of three, namely, the thumb and the middle finger. Figure~\ref{fig:controller} shows the controller, which is made of three main components:

\subsubsection*{\textbf{Low-level controller}}

it is a set of P.I.D. force controllers responsible for maintaining a given force at each fingertip. The control signal is the voltage sent to the motor actuating the proximal joint, while the feedback is the tactile readings at the fingertip. We estimate the force at each fingertip by taking the magnitude of the vector obtained by summing up all the normals at the sensor locations weighted by the sensor response.

\begin{figure}
\includegraphics[width=0.7\linewidth]{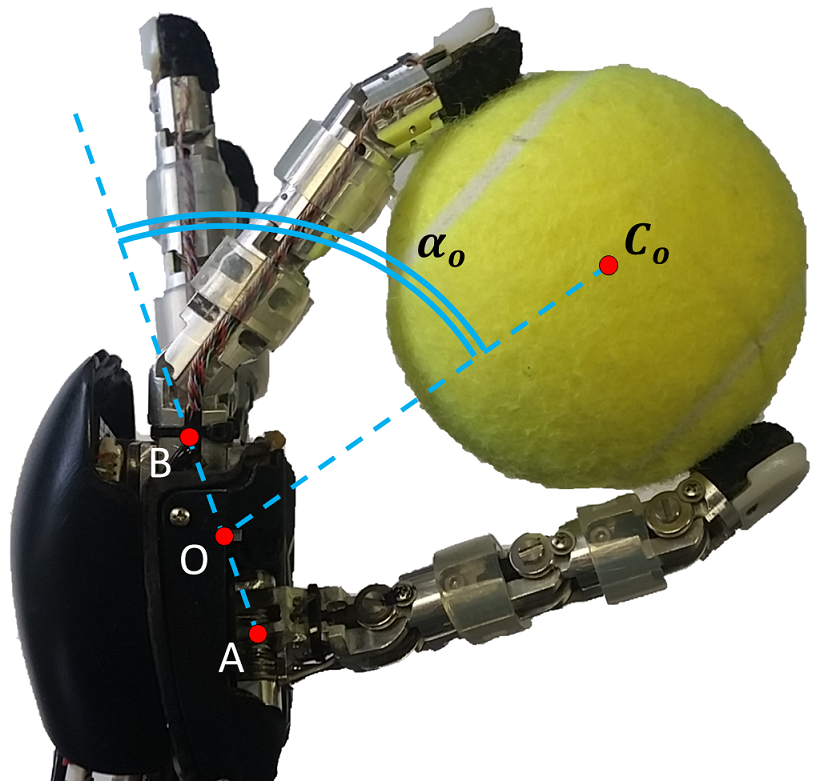}
    \centering
    \caption{The object center $C_{o}$ is defined as the halfway point between the two points of contact. The object position $\alpha_o$ is defined as the angle between the vectors $\vec{OC_{o}}$ and $\vec{OB}$. $A$ and $B$ are set at the base of, respectively, the thumb and the middle finger, while $O$ lies at middle distance between $A$ and $B$.}
    \label{fig:objectPosition}
    
\end{figure}

\subsubsection*{\textbf{High-level controller}}

it is built on top of the low-level force controllers. It stabilizes the grasp by coordinating the fingers to a) control the object position, and b) maintain a given grip strength. The object position $\alpha_o$ is defined as in Fig.~\ref{fig:objectPosition}, and it is controlled using a P.I.D. controller in which the control signals are the set-points of the forces at each finger, while the feedback is the object position error.

The grip strength is the average force applied to the object. It is defined as:

\begin{equation}
g = \frac{f_{th}+f_{mid}}{2}, \\
\label{eqn:gripStrength}
\end{equation}
where $f_{th}$ and $f_{mid}$ are the forces estimated at the thumb and the middle finger, respectively. The target grip strength is maintained by choosing set-points of the forces that satisfy (\ref{eqn:gripStrength}).

\subsubsection*{\textbf{Stable grasp model}}

it is a Gaussian mixture model, trained by demonstration. The robot was presented with stable grasps using objects of different size and shape. The stability of a grasp was determined by visual inspection. A stable grasp is defined as one that avoids non-zero momenta and unstable contacts between the object and the fingertips. We also preferred grasp configurations that are far from joint limits (details are in~\cite{regoli2016hierarchical}). Given the distance, $d$, between the fingers, the model estimates the target object position, $\alpha_o^r$, and the target set of non-proximal joins, $\mathbf{\Theta_{np}}$, to improve grasp stability and make it robust to perturbations. The target $\alpha_o^r$ is used as the set-point of the high-level controller, while the $\mathbf{\Theta_{np}}$ is set directly using a position controller.

\begin{figure}
\includegraphics[width=1.0\linewidth]{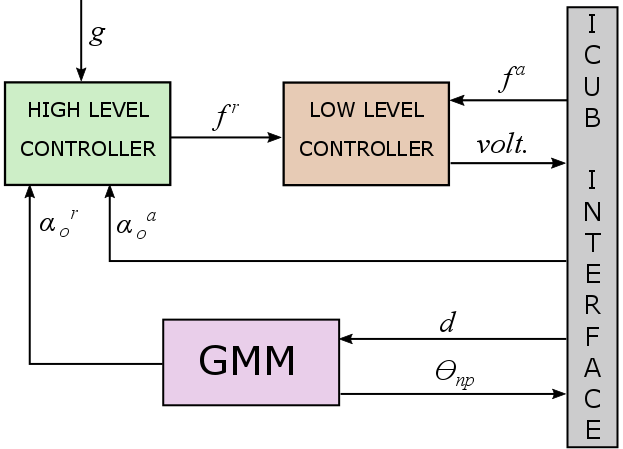}
    \centering
    \caption{Grasp stabilizer control schema. When grasping an object, the distance, $d$, between the points of contact is used by the Gaussian mixture model to compute the reference values of the non-proximal joints, $\mathbf{\Theta_{np}}$, and the object position, $\alpha_o^r$. In order to reach $\alpha_o^r$ and  $g$, the high-level controller sets the appropriate force references, $\mathbf{f^r}$,  of the low-level controller for each finger. The low-level force controller, in turn, sends voltage to the motors actuating the proximal joints to compensate the force error. The actual object position and the actual forces at the fingertips are represented by, respectively, $\alpha_o^a$ and $\mathbf{f^a}$.  }
    
    \label{fig:controller}
\end{figure}

\subsection{The Feature Space}

Once a stable grasp is achieved, the robot manipulates the object to capture its softness and shape by performing two exploratory behaviours: a) squeezing the object between the thumb and the middle finger, and b) wrapping all the fingers around the object. The softness of the object is captured both by the distribution of the forces in the tactile sensor and the deflection of the fingers when the object is squeezed between the fingers of the robot. The shape of the object is captured by wrapping all of the fingers of the robot around it.

As mentioned earlier, the grasp stabilization implies a high degree of repeatability of the achieved pose, independent of the way the object is given to the robot. Thereby, the features produced during the exploratory behaviours exhibit low variance between different grasps of the same object. Which, in turn, increases the accuracy of the classifier.


\subsubsection{Tactile responses}
the distribution of forces in the tactile sensors is affected by the softness of an object. A hard object will exert forces that are strong and concentrated in a local area. A soft object, in contrast, will conform to the shape of the fingertip and exert forces across all tactile sensors. The tactile sensors also capture information on the local shape of the object at the point of contact. We use the tactile responses from the thumb and the middle finger, \tact, in our feature space, since the objects are held between these two fingers.

\subsubsection{Finger encoders}
the finger encoders are affected by the shape and the harness/softness of the object. When the robot squeezes the object, a hard object will deflect the angles of the finger more than a softer object. Since we use only the thumb and the middle finger during the squeezing action, we use both the initial and the final encoder values for these fingers --~\graspInit~and~\graspFin, respectively.

To capture the shape of the object, the robot wraps the rest of its fingers around the object. We also include the encoder data, $\mathbf{\Theta_{wrap}}$,  of these fingers in our feature space.

\subsection{The learning algorithm}
\label{sec:learning}

In order to train the classifier, we used as features the data acquired during the grasping, squeezing and enclosure phase, as described in the previous section. We simply concatenated the collected values, obtaining the feature vector [\graspInit\,\graspFin\, \wrap\,\tact] composed of 45 features, 21 related to the encoders and 24 related to the tactile feedback.

As learning algorithm we adopted Kernel Regularized Least-Squares using the radial basis function kernel. For the implementation we used GURLS \cite{tacchetti2013gurls}, a software library for regression and classification based on the Regularized Least Squares loss function.

\section{Experiments}
\label{sec:experiments}

To test our method, we used the iCub humanoid robot. Its hands have 9 degrees-of-freedom. The palm and the fingertips of the robot are covered with capacitive tactile sensors. Each fingertip consists of 12 taxels~\cite{jamali2015new}.

\subsection{The objects}

We used a set of 30 objects shown in Fig. \ref{fig:confusionMatrices}, of which, 21 were selected from the YCB object and model set~\cite{calli2015benchmarking}. Using a standard set helps in comparing the results of different methods. The objects were selected so that they fit in the iCub robot's hand without exceeding its payload. The YCB object set did not have many soft objects fitting our criteria, hence, we supplemented the set with 9 additional object with variable degree of softness. We also paid attention to choose objects with similar shape but different softness, as well as objects with similar material but different shapes.


\subsection{Data collection}
\label{sec:dataCollectionMethod}

\begin{figure}
\begin{subfigure}[b]{0.22\textwidth}
         \includegraphics[width=\textwidth]{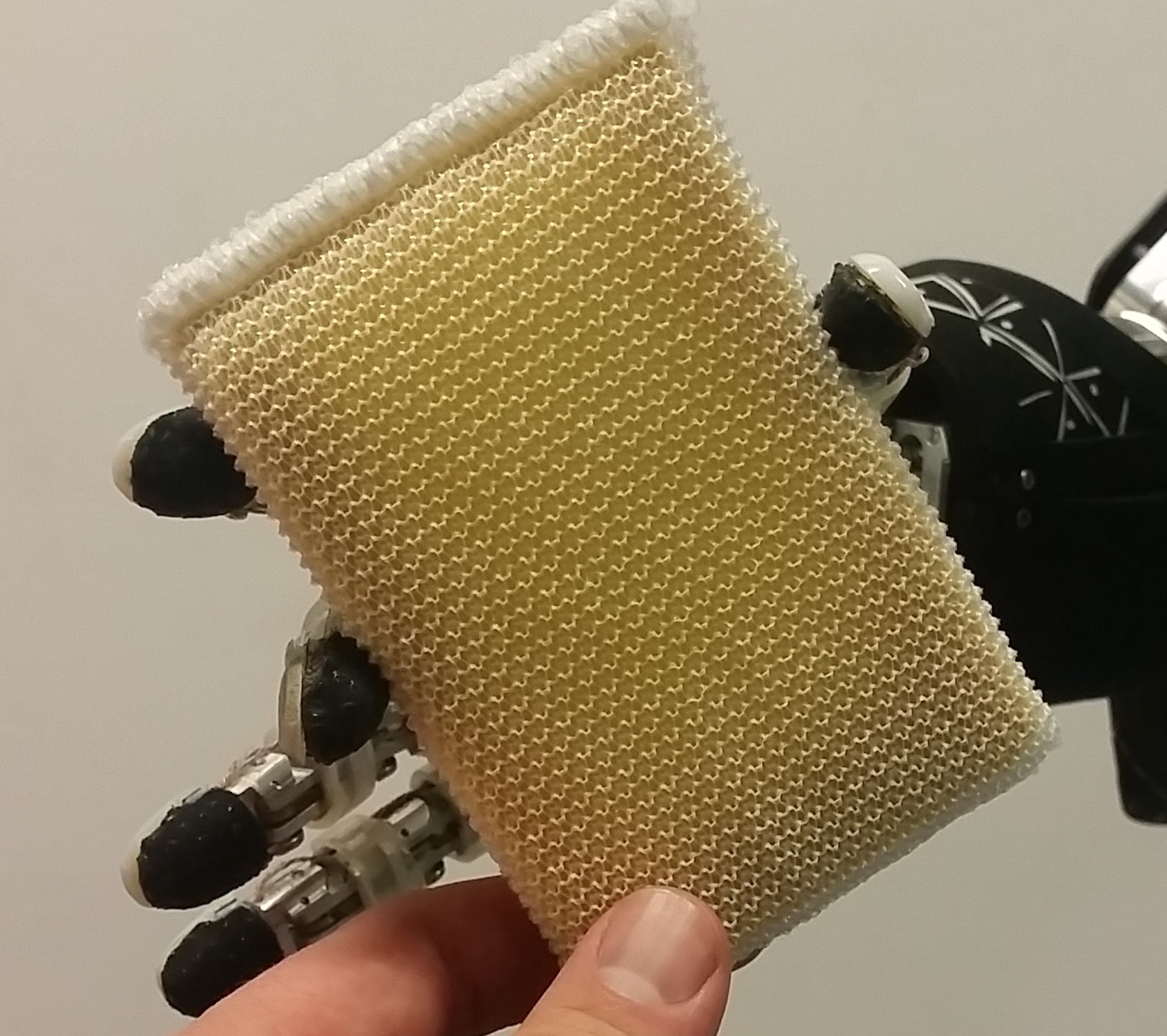}
         \caption{}
         \label{fig:taskStep1}
     \end{subfigure}
     \begin{subfigure}[b]{0.22\textwidth}
         \includegraphics[width=\textwidth]{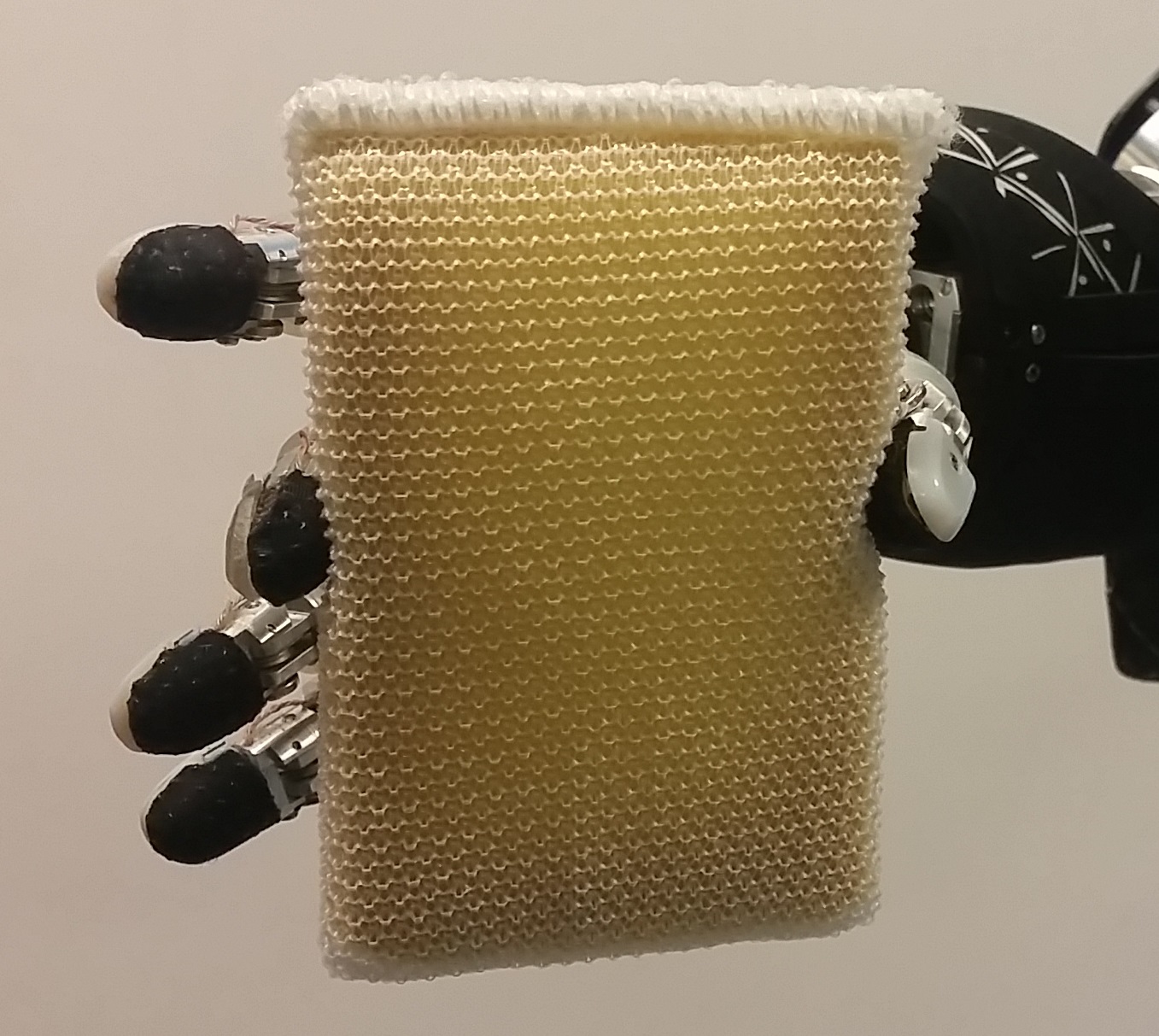}
         \caption{}
         \label{fig:taskStep2}
     \end{subfigure}\\
     \begin{subfigure}[b]{0.22\textwidth}
         \includegraphics[width=\textwidth]{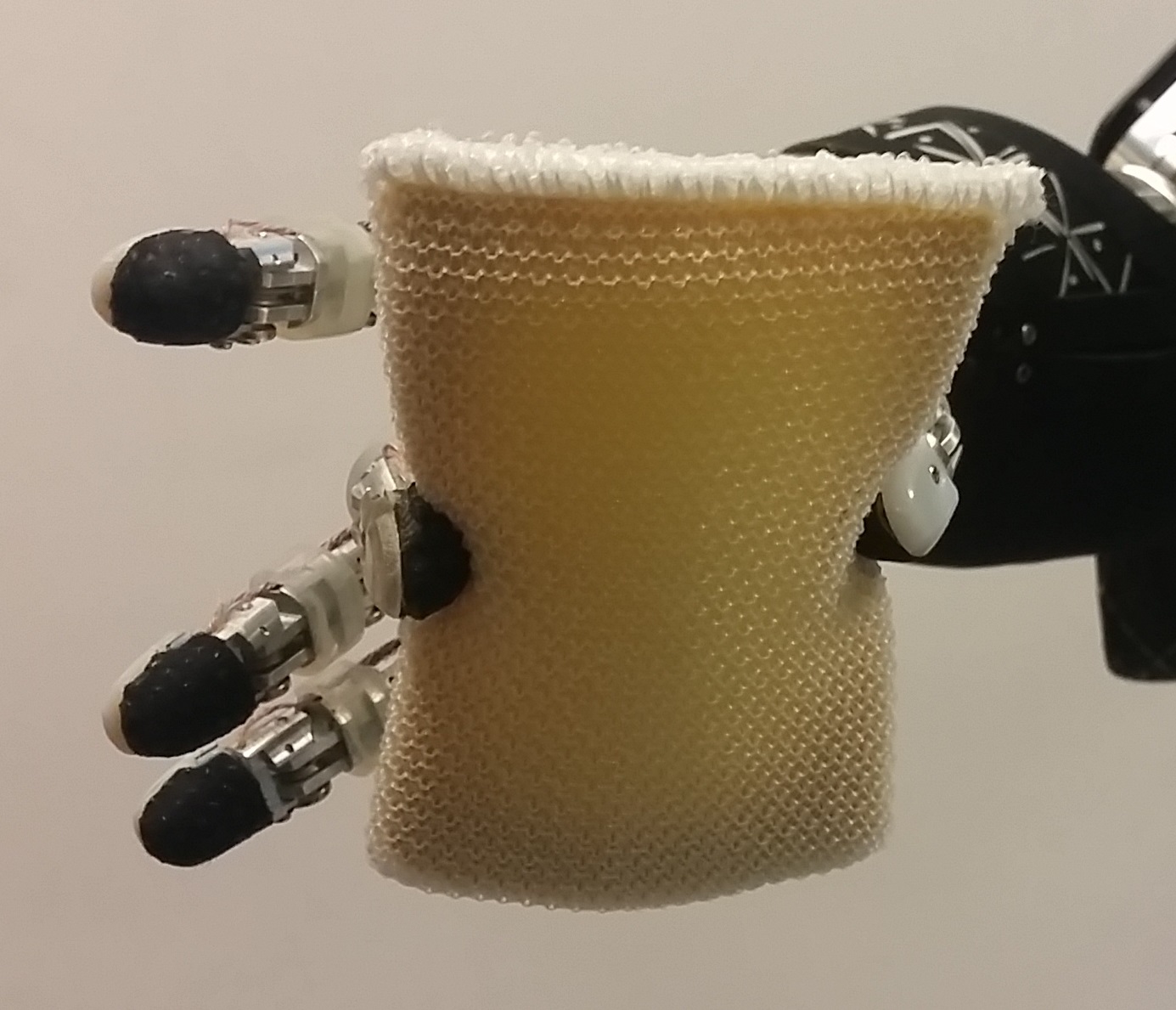}
         \caption{}
         \label{fig:taskStep3}
     \end{subfigure}
         \begin{subfigure}[b]{0.22\textwidth}
         \includegraphics[width=\textwidth]{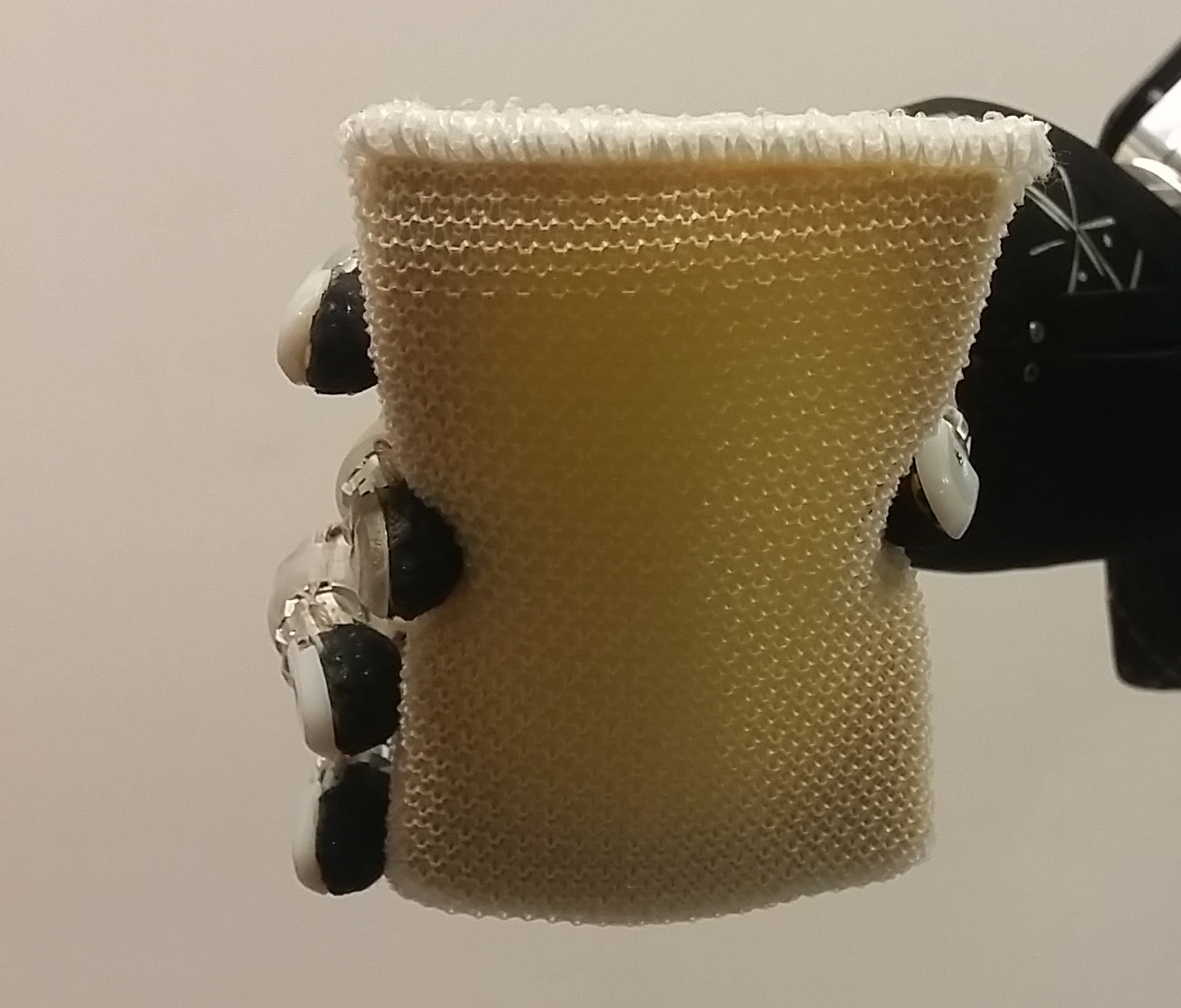}
         \caption{}
         \label{fig:taskStep4}
     \end{subfigure}

\caption{The steps accomplished to identify the object: approach (a), stabilization (b), squeezing (c) and enclosure (d). \lorenzo{Notice that the controller repositions the object irrespective of its initial pose. As discussed in the text this greatly improve repeatability and, consequently, recognition performance.}}
\label{fig:taskSteps}
\end{figure}

The dataset to test our method was collected using the following procedure (depicted in Fig.~\ref{fig:taskSteps}):

\begin{enumerate}
\item The iCub robot opens all of its fingers. 

\item An object is put between the thumb and the middle finger of the robot. The robot starts the approach phase, which consists of closing the thumb and the middle finger until a contact is detected in both fingers. A finger is considered to be in contact with an object when the force estimated at its fingertip exceeds a given threshold. To capture variations in the position and the orientation of the object, each time the object is given to the robot, it is given in a different position and orientation.

\item \label{it:stabilize} At this point the grasp stabilizer is triggered with a given grip strength. The initial value of the grip strength is chosen as the minimum grip strength needed to hold all the objects in the set. The method described in section~\ref{sec:stabilization} is used to improve the grasp stability. When the grasp has been stabilized, the robot stores the initial values of the encoders of the thumb and the middle finger.

\item Then the robot increases the grip strength to squeeze the object and waits for 3 seconds before collecting the tactile data for the thumb and the middle finger. At this point the robot also records the encoder values for the thumb and the middle finger.

\item Finally, the robot closes all of the remaining fingers around the object until all fingers  are in contact with the object. At this point, the robot collects the values of the encoders of the fingers.

\end{enumerate}

These steps were repeated 20 times for each object. To test our algorithm we use a fourfold cross-validation. That is, we divide the dataset into 4 sets. We hold one of the sets for testing and use the other three to train a classifier. This is repeated for all 4 sets. We compute the accuracy and the standard deviation of our classifier using the results of these 4 learned classifiers. 



\subsection{Benchmark experiment}

To test our hypothesis that reaching a stable pose improves the classification results we carried out an experiment in which we disable part of the grasp stabilization. As described earlier and depicted in Fig.~\ref{fig:controller}, the grasp stabilization consists of three modules: the low-level force controller, the high-level controller and the stable grasp model. We only disable the stable grasp model. The other two components are needed to stop the object from slipping and to control the grip strength.

The stable grasp model produces two outputs: a) the target object position, $\alpha_o^r$,  and the target set of non-proximal joints, $\mathbf{\Theta_{np}}$. In the benchmark experiment we calculate the value of $\alpha_o^r$ and the $\mathbf{\Theta_{np}}$ when the thumb and the middle finger make contact with the object. That is, the alpha is set to the current position of the object and the theta is set to the current joint configuration. Apart from this difference, the high-level controller and the low-level force controller are still active, controlling grip strength and maintaining a stable grasp. However, without the stable grasp model, the grasp is less robust to perturbations. 

Henceforth, unless stated otherwise, when we mention that the grasp stabilization is disabled, we mean that we only disable the repositioning based on the GMM. Hence, we collected the data for the benchmark experiment following the same steps as described in section~\ref{sec:dataCollectionMethod} where the grasp stabilization was disabled.

 



\section{Results}
\label{sec:results}

In this section we present the results of our method and show how each of the selected features in our feature space helps in capturing different properties of the objects, namely, the softness/harness and the shape  of the object. This will be followed by a comparison between our method and the benchmark method in which the grasp stabilization is disabled. When reporting the results for brevity we concatenated some of the features: \thetaGrasp = [\graspInit\,\graspFin], and \thetaAll~=~[\graspInit\,\graspFin\,\wrap].

\begin{table}[t!]
    \centering
    \begin{tabular}{cccccc}
      \toprule
					Features	&\graspInit 	& \thetaGrasp & \thetaAll 	& \tact & All\\
      \midrule
      Mean	& 80.5\% 			&93.3\% 				&96.3\%			&95.0\%     &99.0\% \\
      Std	& 2.0\% 			&0.8\% 				&0.7\%			&0.8\%     &0.6\% \\

      \bottomrule
    \end{tabular}
\caption{Classification accuracies using our method with classifiers trained using different set of features.}
    \label{tab:methodAccuracy}
\end{table}

\subsection{Finger encoders}

\begin{figure*}
\centering
\begin{subfigure}[b]{0.43\textwidth}
         \includegraphics[width=\textwidth]{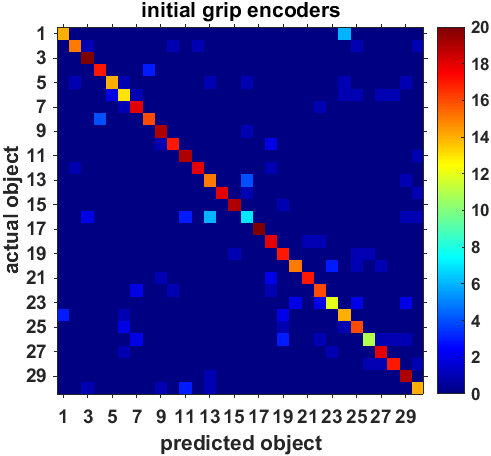}
				 \centering
         \label{fig:confMat1}
     \end{subfigure}
		 \hspace{0.5cm}
     \begin{subfigure}[b]{0.43\textwidth}
         \includegraphics[width=\textwidth]{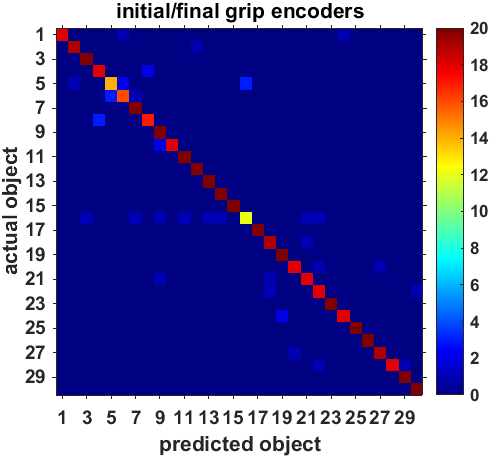}
				 \centering
         \label{fig:confMat2}
     \end{subfigure}\\
     \begin{subfigure}[b]{0.43\textwidth}
         \includegraphics[width=\textwidth]{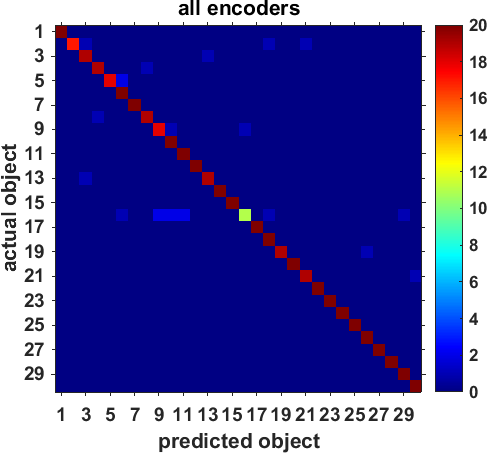}
				 \centering
         \label{fig:confMat3}
     \end{subfigure}
				 \hspace{0.5cm}
         \begin{subfigure}[b]{0.43\textwidth}
         \includegraphics[width=\textwidth]{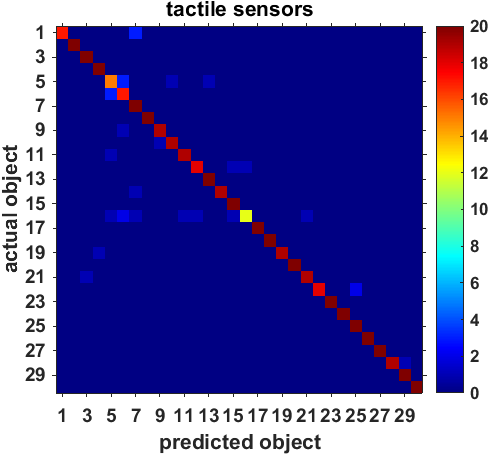}
				 \centering
         \label{fig:confMat4}
     \end{subfigure}
         \begin{subfigure}[b]{0.43\textwidth}
         \includegraphics[width=\textwidth]{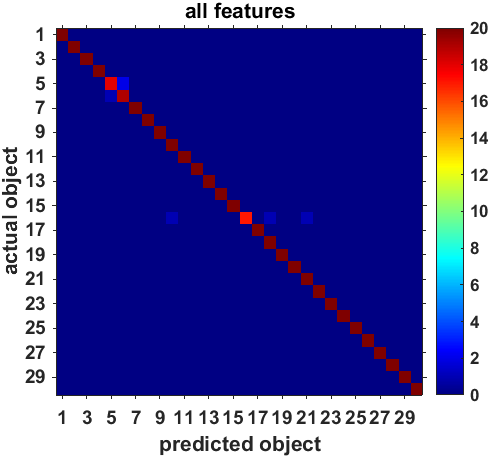}
				 \centering
         \label{fig:confMat5}
     \end{subfigure}
			\hspace{0.5cm}
	     \begin{subfigure}[t]{0.43\textwidth}
	     \vspace{-6.7cm}
         \includegraphics[width=\textwidth]{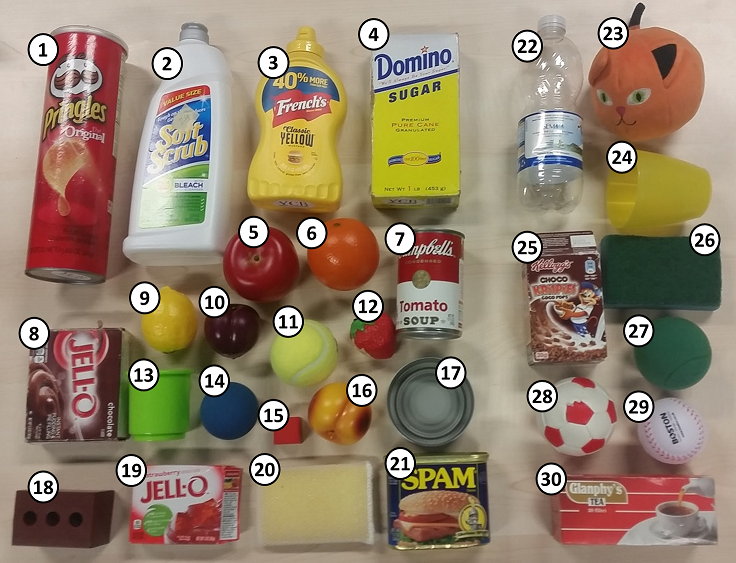}
         \vspace{0.2cm}
         \label{fig:confMatObjects}
     \end{subfigure}

\caption{The confusion matrices obtained using our method with different sets of features. At the bottom right, is the object set used for the experiments. It is composed of 21 objects taken from the YCB object set (left), and additional 9 objects of various degree of softness (right).}
\label{fig:confusionMatrices}
\end{figure*}

To study the effectiveness of the encoder features, we trained a model using different combinations of these 
features. Table~\ref{tab:methodAccuracy} reports the results of these experiments. We notice that using only the initial encoder values, the accuracy is already quite high, 80.5\% $\pm$ 2.0\%, while including the final encoder values of the thumb and the middle finger after squeezing it increases to 93.3\% $\pm$ 0.8\%. This is because the fingers will move considerably if the object is soft, thereby, capturing the softness of the object. Figure~\ref{fig:confusionMatrices} shows the confusion matrices for the experiments. We notice that several pairs of objects such as the tennis ball (11) and the tea box (30) or the sponge (26) and the soccer ball (28) are sometimes confused if only the initial encoders values are used as features, while they are discriminated after the squeezing action.

Finally we analysed the results of including all encoder data, that is, including the data when the robot wraps its fingers around the object. This improved the classification accuracy to 96.3\% $\pm$ 0.7\%. From the confusion matrices we notice that adding such features resolves a few ambiguities, such as the one between the soccer ball (28) and the water bottle (22) and the one between the yellow cup (24) and the strawberry Jello box (19). Indeed, these pairs of objects have similar distance between the points of contact when grasped, and cause similar deflections of the fingers when squeezed, but have different shapes.


\subsection{Tactile responses}

\begin{figure}
\includegraphics[width=1.0\linewidth]{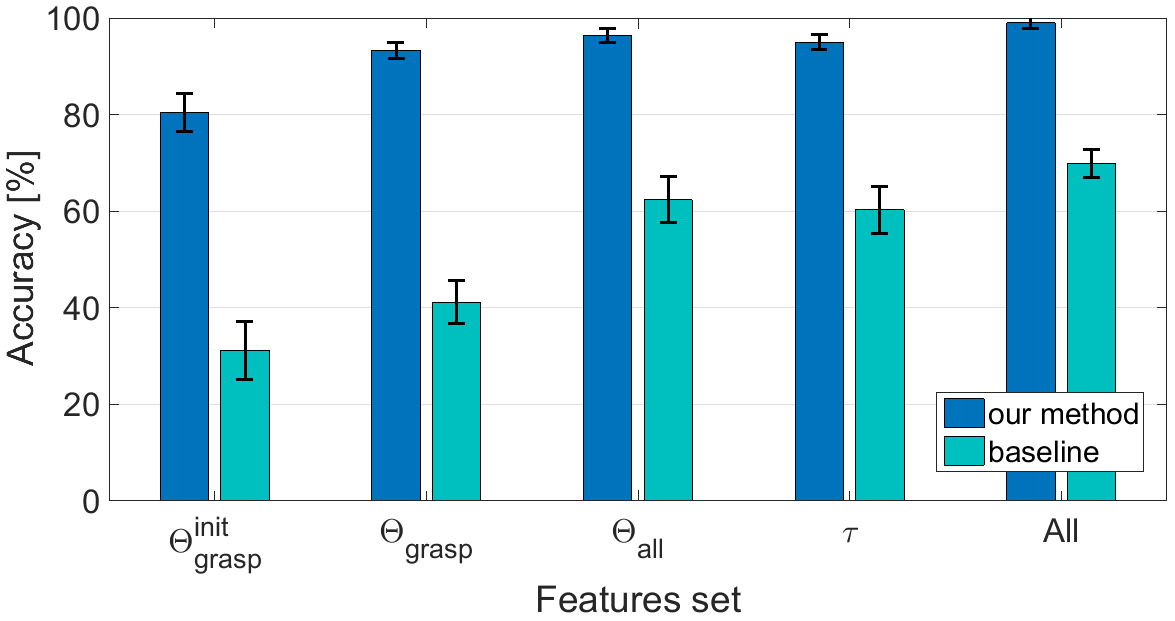}
    \centering
    \caption{Summary of the results comparing our method with the benchmark method for different set of features. It shows that our method outperforms the benchmark method with statistical significance. The error bars are standard deviations.}
    \label{fig:accuracyBars}
\end{figure}

As discussed earlier the tactile sensors are useful in capturing the softness of the objects as well as the local shape of the objects. In Fig.~\ref{fig:accuracyBars} we can see that using only the tactile feedback we get an accuracy of 95.0\% $\pm$ 0.8\%, which is comparable with the 96.3\% $\pm$ 0.7\% obtained using the encoder values. Although they have similar classification accuracy, studying the confusion matrices reveals that objects confused by them are different. For example, the classifier trained using only the tactile data often confuses the Pringles~can~(1) and the tomato~can~(7), since they are hard and share similar local shape. Conversely, due to their slightly different size they are always distinguished by the classifier trained using only encoder data. This means that combining the two feature spaces can further improve the accuracy of the learned classifiers.

\subsection{Combining the two features}

Finally, using the complete feature vector we get an accuracy of 99.0\% $\pm$ 0.6\%. We also notice that the standard deviation in our experiments is decreasing as we add more features. From the confusion matrix we can see that several ambiguities characterizing each individual classifier are now solved. A few objects are still confused due to their similar shape and softness, namely the apple (5) and the orange (6), and the apricot (16) and the prune (10). Less intuitively, the classifier once confuses the apricot with the SPAM can (21), and once it confuses the apricot with the brown block (18).
To explain the confusion between these objects, we notice that there is a particular way to grasp them such that the joints configuration is very similar. \lorenzo{This happens when the middle finger touches the flat side of the apricot, and the little finger misses both objects}.



\subsection{Comparison with the benchmark experiment}

Figure~\ref{fig:accuracyBars} shows the results of running the same analysis on the data collected in the benchmark experiment where the grasp stabilization was removed. The results show that the proposed method performs significantly better than the benchmark experiment, achieving 99.0\% $\pm$ 0.6\%, compared to the benchmark experiment which achieved an accuracy of 69.9\% $\pm$ 1.4\%. This is because the stabilization method proposed in this paper increases the repeatability of the exploration, which makes the feature space more stable. \lorenzo{Indeed, the initial position of the object in the hand strongly affects the collected tactile and encoders data. This variability is reduced using the grasp stability controller.} Note that the accuracy of the benchmark experiment increases as more features are added, showing that the feature space is able to capture the object properties.

\begin{figure}
\includegraphics[width=1.0\linewidth]{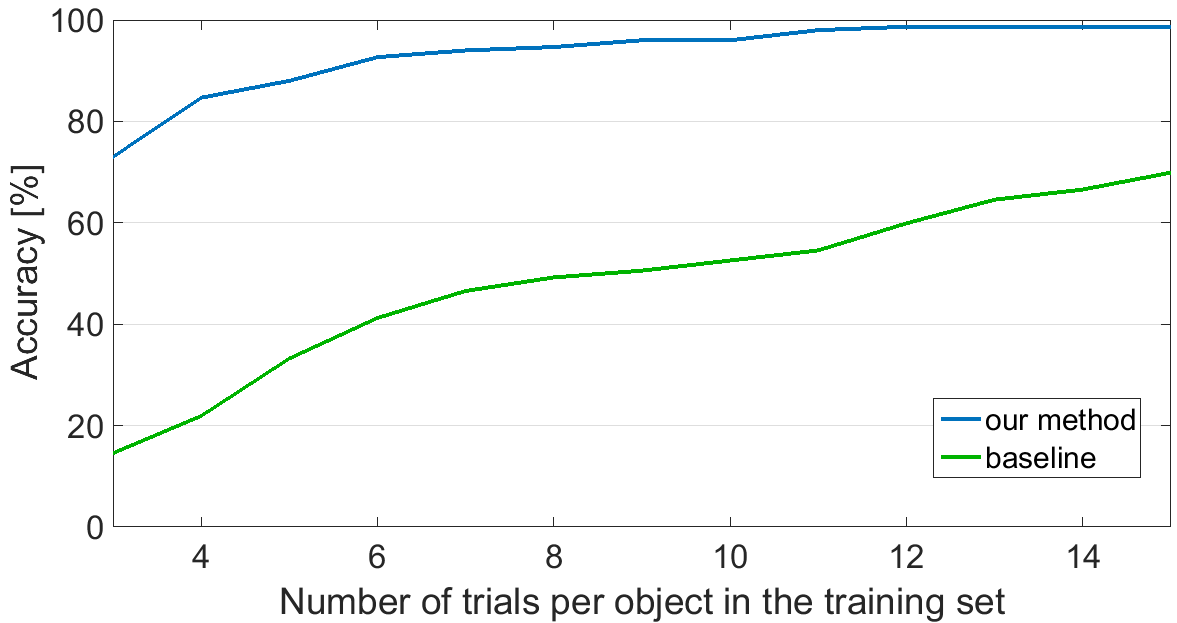}
    \centering
    \caption{The accuracy of our method and the benchmark method as a function of the number of training set samples. Our method obtains high accuracy even with a much lower number of training samples.}
    \label{fig:accuracyTrends}
\end{figure}

We run a further analysis to study the effect of increasing the number of trials in the training set. In this case we always trained the classifier with the complete feature vector and considered 5 \lorenzo{trials} per object for the test set, while we varied the number of trials in the training set between 3 and 15. Figure~\ref{fig:accuracyTrends} shows the results of this analysis. The results show that the proposed method boosts the accuracy of the classification, requiring less samples to be able to distinguish the objects. The trend of the accuracy obtained using the benchmark method suggests that it may improve by increasing the number of samples in the training set. However, this is not preferred because it makes it impractical to collect data on large sets of objects, adversely affecting the scalability of the learned classifier.




\subsection{Results using objects form the YCB set only}
\begin{table}[t!]
    \centering
    \begin{tabular}{cccccc}
      \toprule
									Features	&\graspInit 	& \thetaGrasp & \thetaAll 	& \tact & All\\
      \midrule
      Mean	& 85.0\%             &91.4\%                 &95.0\%            &94.1\%     &97.6\% \\
      Std	& 3.1\%             &1.5\%                 &1.8\%            &1.6\%     &0.5\% \\
      \bottomrule
    \end{tabular}
\caption{Classification accuracies using our method on the YCB objects only. }
    \label{tab:ycbResults}
\end{table}

Finally, in table \ref{tab:ycbResults} we provide the results of our method using only the object from the YCB object set, in order to let researchers having the same dataset compare their results with ours.

\section{Conclusions}
\label{sec:conclusions}

In this work we proposed a method for in-hand object recognition that makes use of a grasp stabilizer and two exploratory behaviours: squeezing and wrapping the fingers around the object. The grasp stabilizer plays two important roles: a) it prevents the object from slipping and facilitates the application of exploratory behaviours, and b) it moves the object to a more stable position in a repeatable way, which makes the learning algorithm more robust to the way in which the robot grasps the object. \lorenzo{We demonstrate with a dataset of 30 objects and the iCub humanoid robot that the proposed approach leads to a remarkable recognition accuracy ($99.0\%\,\pm\,0.6\%$), with a significant improvement of $29\%$ with respect to the benchmark, in which the grasp stabilizer is not used.}

\lorenzo{This work demonstrates that a reliable exploration strategy (e.g. squeezing and re-grasping) is fundamental to acquiring structured sensory data and improve object perception. In future work we will employ an even larger set of objects and explore the use of other control strategies and sensory modalities.}


\bibliographystyle{IEEEtran}
\bibliography{tactileObjectRecognitionBibliography}

\end{document}